# Future Intelligent Autonomous Robots, Ethical by Design. Learning from Autonomous Cars Ethics

**Gordana Dodig-Crnknovic[1,2*], Tobias Holstein[2], Patrizio Pelliccione[3,4]**

[1] Interaction Design Unit, Department of Computer Science and Engineering, Chalmers | University of Technology Gothenburg, Sweden

[2] Division of Computer Science and Software Engineering, School of Innovation, Design and Engineering, Mälardalen University, Västerås, Sweden

[3] Software Engineering Division, Department of Computer Science and Engineering, Chalmers | University of Gothenburg, Gothenburg, Sweden

[4] Department of Information Engineering, Computer Science and Mathematics, University of L'Aquila, L'Aquila, Italy

**\* Correspondence:**
Gordana Dodig-Crnkovic
dodig@chalmers.se

## Abstract

Development of the intelligent autonomous robot technology presupposes its anticipated beneficial effect on the individuals and societies. In the case of such disruptive emergent technology, not only questions of how to build, but also why to build and with what consequences are important. Thus, questions of values, good and bad, right and wrong, must be addressed in the context of the development, implementation, testing, use and disposal of technology. The field of ethics of intelligent autonomous robotic cars is a good example of research with actionable practical value, where a variety of stakeholders, including the legal system and other societal and governmental actors, as well as companies and businesses, collaborate bringing about shared view of ethics and societal aspects of technology. It could be used as a starting platform for the approaches to intelligent autonomous robot development considering human-machine interfaces in different phases of the life cycle of technology. Drawing from our work on ethics of autonomous intelligent robocars, and the existing literature on ethics of robotics, our contribution consists of a set of values and ethical principles with identified challenges and proposed approaches for meeting them. This may help stakeholders in the field of intelligent autonomous robotics to connect ethical principles with their applications. Our recommendations of ethical requirements for autonomous cars can be used for other types of intelligent autonomous robots, with the caveat for social robots that require more research regarding interactions with the users. We emphasize that existing ethical frameworks need to be applied in a context-sensitive way, by assessments in interdisciplinary, multi-competent teams through multi-criteria analysis. Furthermore, we argue for the need of a continuous development of ethical principles, guidelines, and regulations, informed by the progress of technologies and involving relevant stakeholders.

## 1    INTRODUCTION

**Expectations from the Future Intelligent Robots**

"As a game-changing technology, robotics naturally will create ripple effects through society. Some of them may become tsunamis. So it's no surprise that 'robot ethics'—the study of these effects on ethics, law, and policy—has caught the attention of governments, industry, and the broader society, especially in the past several years." (Lin, Abney, and Bekey 2011).

Thus an important question to answer is: "What should we want from a robot ethic?" asked by (Asaro 2020) The suggested answer emphasizes gradual development of robots so that they successively acquire greater ethical capabilities. As the "the overarching interest in robot ethics ought to be the practical one of preventing robots from doing harm", responsibility assignment in complex socio-technical should be regulated by legal theory, according to Asaro.

Apart from the robot ethics, with ethical problems resulting from the use of robots, Vincent Müller (Müller 2020) mentions the issue of "ethical status of the robots themselves" and "machine ethics" as the effort to make robots ethical. Some authors (Wallach and Allen 2009; Operto 2011; Dodig-Crnkovic and Çürüklü 2012) propose building robots "ethical by design".

Further key references on the ethics of intelligent robotics include (Tzafestas 2015; Capurro 2009; Asaro 2006; Veruggio and Operto 2008). While providing an overview, Capurro focusses on ethics and robotics from an intercultural perspective. Veruggio and Operto define ethical and social issues of robotics in their chapter in the *Springer Handbook of Robotics*, whereas Asaro asks an important question about the role of a robot ethic. We share his view concerning three topics: "the ethical systems built into robots, the ethics of people who design and use robots, and the ethics of how people treat robots." Thus, we fulfill those by treating robots as socio-technical systems.

## 2    Lessons Learned from Autonomous Cars Ethics.
##      Ethical Analysis with Requirements, Challenges and Possible Approaches

This mini-review is based on findings from our book chapter (Holstein, Dodig-Crnkovic, and Pelliccione 2021) on ethical and social aspects of self-driving cars, robots classified as "mobile service robots" (Ben-Ari et al. 2018). Those are vehicles capable of perceiving their environment and driving without (or with little) human intervention. They combine advanced sensing, controlling, and artificial intelligence with autonomous safety-critical decision making. Ethical aspects of autonomous cars (autonomous vehicles, driverless cars, automated cars or robocars) have lately got a lot of attention from the general public, ethicists, researchers, industry and decision-makers.

The chapter (Holstein, Dodig-Crnkovic, and Pelliccione 2021) is based on the literature on value-based design and current guidelines (Friedman and Kahn Jr. 2003; Friedman et al. 2013; The IEEE Global Initiative on Ethics of Autonomous and Intelligent Systems 2019; European Group on Ethics in Science and New Technologies (EGE) 2018; High-Level Expert Group on Artificial Intelligence (AI HLEG) 2019; Floridi et al. 2018), extracted the list of topics/requirements/values of relevance for real-world automated/self-driving cars that we presented in a table, for both technical and social ethical challenges. A similar convergence of ethics requirements has been observed in AI-ethics literature globally (Morley et al. 2020).

In (Holstein, Dodig-Crnkovic, and Pelliccione 2021; 2020; 2018; Holstein and Dodig-Crnkovic 2018), we developed the approach to the practice-oriented, real-world ethics for self-driving cars. We started from the most important technological and societal requirements based on extensive literature studies and identified challenges. With practical applications in mind, we searched for the consensus among international studies about the most important ethical issues as requirements for intelligent cars. After producing such a list (Table 1) we identified challenges and approaches to addressing them. We contacted leading experts in the field, as well as colleagues researching in interaction design, software engineering, and philosophy of technology, and in the dialogue with the general public, as well as through the discussions in the public lectures we tested and concretized our ideas. The details of this work are given in (Holstein, Dodig-Crnkovic, and Pelliccione 2021).

In the next step, we compared the Table 1 developed for robotic cars, with the ethics requirements found in research on robots ethical by design, medical robots, industrial robots, and other intelligent robots to whom we delegate responsibilities (Giovagnoli, Crucitti, and Dodig-Crnkovic 2019; Georgieva and Dodig-Crnkovic 2011; Çürüklü, Dodig-Crnkovic, and Akan 2010; Dodig-Crnkovic and Persson 2008; Dodig-Crnkovic 2009; 2010; Dodig-Crnkovic and Çürüklü 2012). We have also taken into account findings of works of (Lin, Abney, and Bekey 2011) (Tzafestas 2015; Capurro 2009; Asaro 2006; Veruggio and Operto 2008) and (Müller 2020), regarding ethical concerns and requirements for robotics. The recent whitepaper from the UK Robotics and Autonomous Systems Network (Winfield, McDermid, et al. 2019) proposes seven ethical concerns of robotics (Bias, Deception, Employment, Opacity, Safety, Oversight and Privacy). All except for deception are part of our original framework, (Holstein, Dodig-Crnkovic, and Pelliccione 2021; 2020; 2018). Here we added deception as an important concern important for social robotics, and particularly for intelligent robotic companions in case of vulnerable users (Boden et al. 2017).

Ethical aspects of technical and social challenges presented in the following table, modified from self-driving cars to correspond to the case of robotics systems in general, are reproduced from (Holstein, Dodig-Crnkovic, and Pelliccione 2021; 2018),with permission.

*Table 1. Summary of the ethical aspects of intelligent autonomous robots, grouped by requirement*

| A. The ethical aspects of technical challenges | | |
|---|---|---|
| **Requirements** | **Challenges** | **Approaches** |
| Safety | Hardware and software adequacy. Vulnerabilities of machine-learning algorithms. Control of trade-offs between safety and other factors (like economic) in the design, manufacturing and operation. Possibility of intervention in case of major failure of the system and graceful degradation. Systemic solutions to guarantee safety in organizations (regulations, authorities, safety culture). | Setting safety as the first priority. Learning from the history of automation. Learning from experience of use. Specification of how a system will behave in cases when autonomous operation is disabled (safe mode). Preparedness for handling "loss of control" situations- autonomous systems running amok. Regulations, guidelines, standards being developed as the technology develops. |
| Security | Minimal necessary security requirements for deployment of the system. Security in the context and connections. Deployment of software updates. Storing and using received and generated data in a secure way. | Technical solutions to guarantee minimum security under all foreseeable circumstances. Anticipation and prevention of the worst-case scenarios. Accessibility of all data, even in the case of accidents, learning from experience. |
| Privacy | Trade-offs between privacy and data collection/recording and storage/sharing. | Following/applying legal frameworks to protect personal data, such as GDPR. |
| Transparency | Information disclosure, what and to whom. Transparency of algorithmic | Assurance of transparency and insight into decision making. Active sharing of knowledge |
| | decision making. Transparency in the techno-social ecosystem. | to ensure the interoperability of systems and services. |
| Algorithmic Fairness | Algorithmic decision making is required to be fair and not to discriminate on the grounds of race, gender, age, wealth, social status etc. | This requirement is related to transparency of decision making and expectation of explainability of the ground for decision making (e.g., right of explanation is enforced by GDPR as stated in Recital 71 (E.U. 2016)). |
| Reliability | Reliability of hardware, sensors and software and need for redundancy. Reliability of required networks and solution for the case when the network is unavailable. | Definition of different levels for reliability, such as diagnostics, hardware, sensors, software, and external services, set the ground for reliability measures of the system and its components. The standardized process required to shift from fail-safe to fail-operational architecture. |
| Environmental Sustainability | Environmental sustainability ethics refers to new ways of production, use, and recycling for robotic systems. | Production, use, and disposal/recycling of technology rise sustainability issues (materials, processes, energy) that must be addressed. |
| Intelligent Behavior Control | Intelligent behaviour may lead to unpredictable situations resulting from learning and autonomous decision making. | Development of self-explaining capability and other features ensuring desired behavior in intelligent software. |
| Transdisciplinary - Systemic Approach | Ethics in/for/by/through/of design. Requirements engineering, software-hardware development, learning, legal and social aspects, software-hardware interplay. | Adoption of transdisciplinarity and system approaches is increasing and should be strengthened even more. |
| Quality | Quality of components. Quality of decision making. Lifetime and maintenance. QA process. Adherence to ethical principles/guidelines. | Ethical deliberations included in the whole process starting with design and development. Ethics-aware decision making to ensure ethically justified decisions. |

| *B. Ethical aspects of social and individual challenges of intelligent autonomous robots* | | |
|---|---|---|
| **Requirements** | **Challenges** | **Approaches** |
| Non-maleficence | Risk of technology causing harm. Disruptive changes in the labor market. Transformation of related businesses, markets and business models (manufacturers, insurances, etc.). | Partly covered by technical solutions. Preparation of strategic solutions for people losing jobs. Learning from historic parallels to industrialization and automatization. |
| Stakeholders' involvement | Participation of different stakeholders – from professionals in designing, developing, maintenance and recycling, to users, and general public. | Active involvement of stakeholders in the process of design and requirements specification as well as decisions of their use. |
| Beneficence | Establishment of values and priorities: Ensuring that shared public values will be embodied in the technology, with interests of minorities taken into account. | Initiatives as "AI for good" exemplify this expectation that new technologies not only do not cause harm but actively do good for its stakeholders. |
| Responsibility and Accountability | Assignment and distribution of responsibility and accountability as some of central regulative mechanisms for the development of new technology. They should follow ethical principles. | The Accountability, Responsibility and Transparency (ART) principle based on a Design for Values approach includes human values and ethical principles in the design processes (Dignum 2019). |
| Freedom and Autonomy | Freedom of choice for a human hindered or disabled by the system. | The freedom of choice determined by regulations. Determination and communication of the amount of control that humans have. |
| Social Sustainability | In the domain of business, social sustainability is about identifying and managing business impacts on people. In the case of social robotics, the impact of social robots on society is central. | Pursuing social equity, community development, social support, human rights, labour rights, social responsibility, social justice, etc. |
| Social Fairness | Ascertaining fairness of the socio-technological system. | Fairness of the decision-making. Related to inclusiveness, transparency and explainability. |
| Dignity and Solidarity | Challenges come from the lack of common wholistic view. | This requirement should apply to the entire socio-technological system. |
| Social Trust | Establishing trust between humans and robots as well as within the social system involving robots. | Further research on how to implement trust across multiple systems. Provision of trusted connections between components as well as external services. |
| Justice: legislation, standards, norms, policies, and guidelines | Keeping legislation up to date with the current level of technology, and proactively meeting emergent developments. Creating and defining global legislation frameworks. Including ethical guidelines in design and development processes. | Legislative support and contribution to global frameworks. Ethics training for involved stakeholders. Establishment and maintenance of a functioning socio-technological system in addition to functional safety standards. |
| Cognitive and psychological effects of social/ companion robots on humans | Personal integrity Cognitive load Deception | Further research on how social robots and especially increasingly intelligent and human-like robot companions affect users. Solid understanding of effects, after stakeholders' interests are taken into consideration should be followed by regulation/legislation. Humanoid or zoomorphic robots may cause emotional attachment to some users. "Robots should not be designed in a deceptive way to exploit vulnerable users" (Boden et al. 2017) |

Comparison of the above Table 1 with the one in (Holstein, Dodig-Crnkovic, and Pelliccione 2021) shows that self-driving robotic cars provide a good baseline of ethics requirements, since they have almost all of the requirements that apply to other autonomous intelligent robotic systems. However, cognitive and psychological effects of social/ companion robots on humans require specific analysis that is outside of the scope of autonomous cars ethics.

## 3    Comparative Analysis of the Application of the Autonomous Cars Ethics Framework Extended to other Classes of Intelligent Robots

In order to present major ethical challenges for the development of future intelligent autonomous robotic systems, we have chosen the following classes of robots, representative of different kinds of ethical aspects, and compared their ethical challenges with our approach developed for autonomous robotic cars. We cover intelligent autonomous robots used in the following applications: transport, social, industrial, medical, military and entertainment.

The three authors of this mini review ranked the different requirements independently to see similarities and differences in assessment, which resulted in Figure 1. Among the authors requirements are regarded as more or less important for a certain robot class, depending on the perspective and experiences of the assessor. Obviously, ethical aspects for any class of robots will require the evaluation by interdisciplinary, multi-competent teams through multi-criteria analysis. Applications of the framework, priorities, and actual processes are *context-sensitive and must be decided for a given application*. They will evolve over time and must keep pace with technology development. For example Waymo has established an interdisciplinary team called the Waymo Safety Board (executive leaders from Safety, Engineering and Product teams) to maintain and improve their safety (Webb et al. 2020).

Applying an ethical approach such as ours for bridging "from principles to practice" gap, (Morley et al. 2020) points out in 'A way forward' section that "there is a need for a more coordinated effort, from multi-disciplinary researchers, innovators, policymakers, citizens, developers and designers, to create and evaluate new tools and methodologies, in order to ensure that there is a 'how' for every 'what'". That is exactly what is needed at this stage of development.

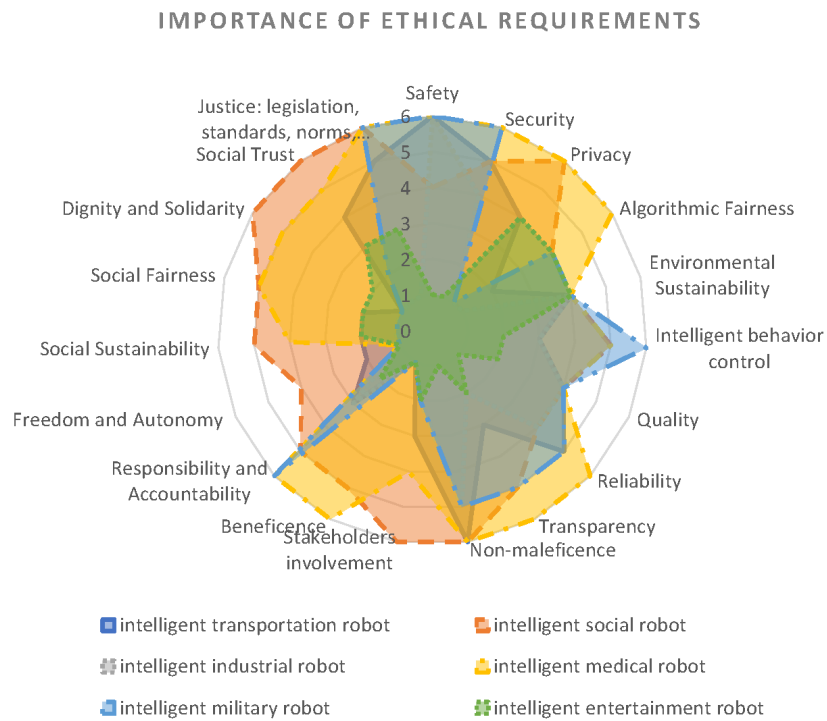

Figure 1. Spider web diagram illustrating the importance of different ethical requirements in different types of robots based on the independent ranking by the three authors

## 4 Context-specific Application of the Proposed Ethical Approach for Intelligent Robotic Systems

As (Morley et al. 2020) argue, we need not only principles but also the connection to the practices. Our Table 1 summarizes ethical challenges related to technical and social aspects of robotics systems. We distinguish between three contexts in which robots appear: 1) the context of application, 2) the context of design/production/maintenance, and 3) the context of oversight.

1) The context of application

In a systemic view of emergent technology, we cannot analyze software in separation from hardware, and we cannot disconnect engineering from human, social, and organizational factors (Leveson 2011; 2020). We have shown how this applies to self-driving cars. The way robots interact with the real-world can be presented in the same abstract phases of "sense, decide, act", which form an iterative process, supported and extended by AI. The context and perspective towards an emergent technology sets the boundaries for each challenge in each of the phases. E.g., privacy for a robot that does not "sense" (with limited perception, recording, etc.) or due to limited context of use (e.g., industrial robot) is less important than a privacy for a self-driving car (more capabilities to sense/ record information and a less restricted context of use). Privacy for self-driving cars is less important than privacy for social robots which relate closely to humans and possess capabilities of sensing, recording, local processing and transmitting information.

2) The context of design/production/maintenance

For different types of robots, a detailed analysis and concretization of design/production/maintenance for their specific contexts must be done. (Spiekermann 2015) presents the value-based view of ethical IT innovation, while (Morley et al. 2020) provide useful advice helping to bridge the conceptual gap between ethical principles and engineering practice. What we found to be most important in this context is collaboration, communication, systems thinking, and solving real-world problems *guided by the needs of stakeholders*. Globally, priorities and values are constantly changing and negotiating in light of new developments. Ethical aspects are present in the technology from research to practice, design, development, implementation, testing, verification and management of the entire system lifecycle, in the iterative process of continuous improvement (Leveson 2011; 2020; Holstein, Dodig-Crnkovic, and Pelliccione 2021; 2018; Dodig-Crnkovic and Çürüklü 2012). Engineers do concretization of the steps for every requirement for each specific context. Frameworks provide guidelines for the approach and require contextual domain knowledge for the specific application. E.g., safety requirements for medical robots are clearly different from the safety requirements for military robots or nano-robots which are examples of robotic systems that pose new ethical challenges because of their microscopic size, the ability of self-replication, mutation and possibility to escape out of control.

3) The context of oversight

Finally, it is important to point out the importance of the oversight of "the total ecology of the socio-technological system, where ethics is ensured through education, constant information sharing and negotiation of priorities in the value system", (Holstein, Dodig-Crnkovic, and Pelliccione 2021). The development of technologies, is followed by the interest of the public and other stakeholders, which is followed by legislation, rules and guidelines (Boden et al. 2017; Winfield, Michael, et al. 2019). As argued in (Dodig-Crnkovic and Çürüklü 2012), this process is a recursive socio-technological learning process, where experiences and new developments lead to improved oversight/regulation/legislation. It is necessary to ensure the transparency of those processes so to enable independent evaluations and efficient learning.
Applying an ethical framework, such as ours or (Floridi et al. 2018) for a good technology-based society, in complex and often unforeseen circumstances of real-world applications, it is important to know how to interpret such general requirements to make ethical choices under uncertainty (Dennis et al. 2014).

## 5    Conclusions

This short review of robotic ethics as a precondition for our trusting the robots with our future, is based on our approach for the ethics of autonomous cars. It argues that ethics of intelligent autonomous robots must permeate application, design/production/maintenance and oversight within their techno-social system by learning from experience (Charisi et al. 2017; Dodig-Crnkovic and Çürüklü 2012; Holstein, Dodig-Crnkovic, and Pelliccione 2021; 2018).

Both (Leveson 2020) and our study emphasize the need for a system-level approach involving *the entire software-hardware system* as well as human, organizational and social levels.

We identified a gap between general principles and practical context-dependent applications when multi-criteria decision making is expected with lists of most important ethical issues to be addressed. The problem is always resolved in its context by a multidisciplinary team with the necessary competences. (Morley et al. 2020).

We argue for the need of a continuous development of ethical principles, guidelines, and analyses, as well as regulatory documents, informed by the progress of technologies and involving all relevant stakeholders.

Ethical deliberation should become a natural part of the techno-social domain. That is the way towards trustworthy intelligent autonomous robots. Still a lot of work remains to be done.

## 6    Conflict of Interest

*The authors declare that the research was conducted in the absence of any commercial or financial relationships that could be construed as a potential conflict of interest.*

## 7    Author Contributions

GDC, TH and PP contributed with the content development, assessment of ethical aspects in different classes of robots, and writing of the manuscript; TH and GDC contributed to bibliography exploration. All authors contributed with perspectives, discussed and revised the manuscript, and approved it for publication.

## 8    Funding

Research partly supported by the Centre of EXcellence on Connected, Geo- Localized, and Cybersecure Vehicle (EX-Emerge), funded by Italian Government under CIPE resolution n. 70/2017.

## 9    Acknowledgments

The authors want to acknowledge the valuable comments and suggestions by Damian Gordon, Mohammad Obaid and two anonymous reviewers.